\documentclass[sigconf, screen]{acmart}

\usepackage{bbm}  
\usepackage{adjustbox}
\usepackage{array}
\usepackage{multirow}
\usepackage{colortbl}
\usepackage{ulem}
\usepackage{makecell}
\usepackage{balance}
\usepackage{pifont}

\definecolor{mygray}{gray}{.9}

\AtBeginDocument{%
  \providecommand\BibTeX{{%
    \normalfont B\kern-0.5em{\scshape i\kern-0.25em b}\kern-0.8em\TeX}}}

\setcopyright{acmcopyright}
\copyrightyear{2022}
\acmYear{2022}
\setcopyright{acmcopyright}\acmConference[MM '22]{Proceedings of the 30th ACM
International Conference on Multimedia}{October 10--14, 2022}{Lisboa, Portugal}
\acmBooktitle{Proceedings of the 30th ACM International Conference on Multimedia
(MM '22), October 10--14, 2022, Lisboa, Portugal}
\acmPrice{15.00}
\acmDOI{10.1145/3503161.3547752}
\acmISBN{978-1-4503-9203-7/22/10}

\begin{document}


\title{Cross-Modality Domain Adaptation for Freespace Detection: A Simple yet Effective Baseline}

\author{Yuanbin Wang}
\affiliation{
  \institution{Institute of Artificial Intelligence, Beihang University}
\country{}
}
\affiliation{
  \institution{Hangzhou Innovation Institute, Beihang University}
\country{}
}
\affiliation{
  \institution{SenseTime Research}
\country{}
}

\author{Leyan Zhu}
\affiliation{
  \institution{Institute of Artificial Intelligence, Beihang University}
\country{}
}
\affiliation{
  \institution{Hangzhou Innovation Institute, Beihang University}
\country{}
}
\affiliation{
  \institution{SenseTime Research}
\country{}
}

\author{Shaofei Huang}
\authornote{Corresponding author.}
\affiliation{
  \institution{Institute of Information Engineering, Chinese Academy of Sciences}
\country{}
}
\affiliation{
  \institution{School of Cyber Security, University of Chinese Academy of Sciences}
\country{}
}

\author{Tianrui Hui}
\affiliation{
  \institution{Institute of Information Engineering, Chinese Academy of Sciences}
\country{}
}
\affiliation{
  \institution{School of Cyber Security, University of Chinese Academy of Sciences}
\country{}
}

\author{Xiaojie Li}
\affiliation{
  \institution{SenseTime Research}
\country{}
}

\author{Fei Wang}
\affiliation{
  \institution{SenseTime Research}
\country{}
}

\author{Si Liu}
\affiliation{
  \institution{Institute of Artificial Intelligence, Beihang University}
\country{}
}
\affiliation{
  \institution{Hangzhou Innovation Institute, Beihang University}
\country{}
}

\renewcommand{\shortauthors}{Yuanbin Wang et al.}

\begin{abstract}
As one of the fundamental functions of autonomous driving system, freespace detection aims at classifying each pixel of the image captured by the camera as drivable or non-drivable. Current works of freespace detection heavily rely on large amount of densely labeled training data for accuracy and robustness, which is time-consuming and laborious to collect and annotate. 
To the best of our knowledge, we are the first work to explore unsupervised domain adaptation for freespace detection to alleviate the data limitation problem with synthetic data. 
We develop a cross-modality domain adaptation framework which exploits both RGB images and surface normal maps generated from depth images.
A Collaborative Cross Guidance (CCG) module is proposed to leverage the context information of one modality to guide the other modality in a cross manner, thus realizing inter-modality intra-domain complement.
To better bridge the domain gap between source domain (synthetic data) and target domain (real-world data), we also propose a Selective Feature Alignment (SFA) module which only aligns the features of consistent foreground area between the two domains, thus realizing inter-domain intra-modality adaptation.
Extensive experiments are conducted by adapting three different synthetic datasets to one
real-world dataset for freespace detection respectively. 
Our method performs closely to fully supervised freespace detection methods ($93.08\%$ \textit{v.s.} $97.50\%$ F1 score) and outperforms other general unsupervised domain adaptation methods for semantic segmentation with large margins, which shows the promising potential of domain adaptation for freespace detection. 
\end{abstract}



\maketitle

\begin{figure*}[!htbp]
  \includegraphics[width=\textwidth]{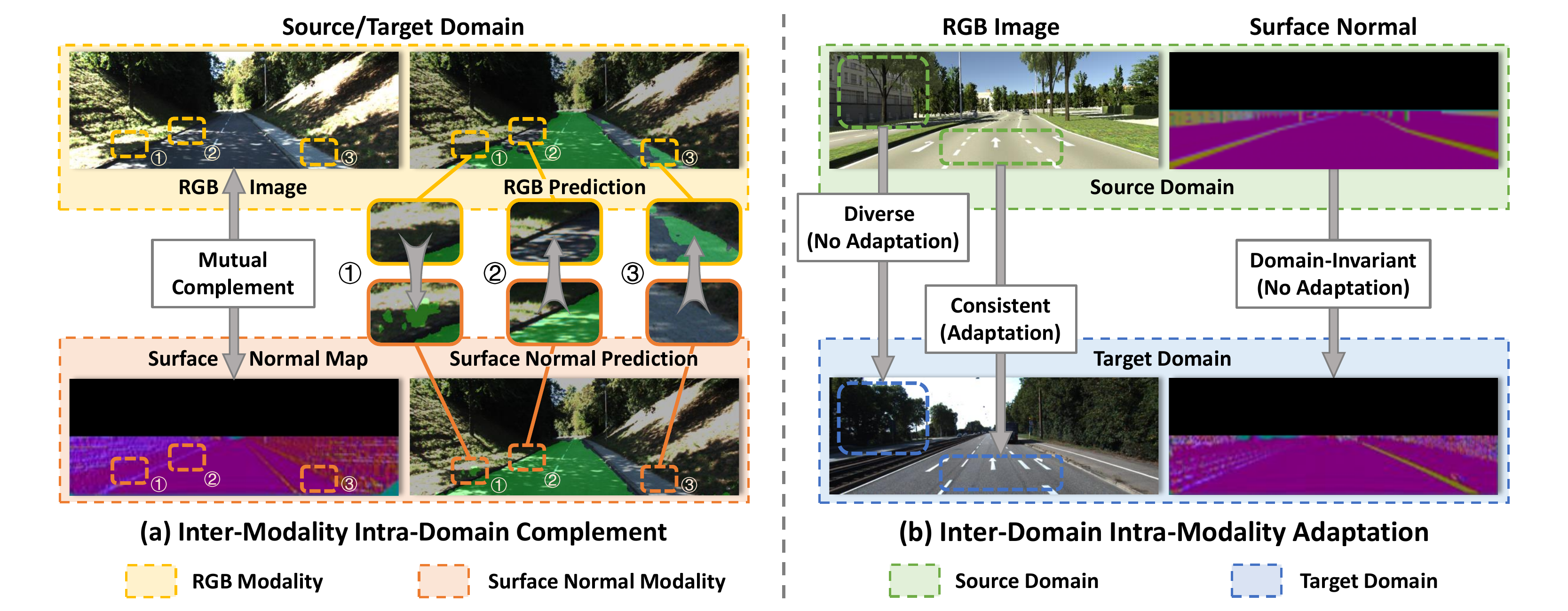}
  \caption{Illustration of our motivation. (a) Inter-modality intra-domain complement. RGB model either mis-classifies road with shadow into non-road area (box\ding{193}) or fails to distinguish between road and sidewalks (box\ding{194}), while surface normal model falsely categorizes grassland into road area (box\ding{192}). These two modalities are complementary to each other. (b) Inter-domain intra-modality adaptation. Foreground areas of RGB images from both domains belong to the same semantic category which is easy to adapt, while background areas are diverse in semantic categories and styles which are difficult to adapt (left column). Surface normal maps from both domains are domain-invariant in style, thus they are not adapted in our method (right column). The top half of surface normal map is visualized in black because of lacking depth information of this region.}
  \label{fig:intro}
\end{figure*}

\section{Introduction}
Autonomous driving~\cite{levinson2011towards, caesar2020nuscenes, sun2020scalability, wang2022keypoint} has drawn remarkable attention in recent years from both industry and academia for pursuing intelligent transportation.
As one of the essential functions of autonomous driving system, freespace detection aims to perform binary classification on each pixel of the image to label it as drivable or non-drivable, so as to realize safety-ensured vehicle navigation.
Current freespace detection methods~\cite{qi2018dynamic, chen2016dynamically, chen2018parallel, lu2019monocular, sun2019reverse, munoz2017deep} usually depend on massive labeled training data which is costly to collect and annotate, thus restricting their practical applications.
Concretely, vehicles equipped with high-definition cameras are required to drive under various scenarios (\textit{e.g.}, weather conditions, seasons, cities and traffics), which is time-consuming and expensive.
It is also laborious to annotate captured road images with pixel-level dense masks.
In order to alleviate this data limitation, synthetic images can actually serve as a large-scale data source with much-reduced collection and annotation costs.
However, there is a huge gap between synthetic and real-world domains~\cite{ghifary2016deep, pizzati2020domain, tsai2018learning}, thus leading to performance drop when directly applying a model trained on synthetic images to real ones.

To bridge the domain gap, we propose to investigate unsupervised domain adaptation for freespace detection (UDA-FD), where only labels of source domain are available.
To the best of our knowledge, the UDA-FD task has not been explored by previous works.
Since RGB data is vulnerable to extreme visual conditions like low illumination, depth data, which could be obtained with stereo cameras, becomes a promising supplement~\cite{fan2020sne, chang2022fast, wang2021sne} due to its good trade-off between reliability and cost.
Since freespace can be hypothesized as a plane, on which the points have similar surface normals, we follow~\cite{fan2020sne} to extract surface normal maps from depth images as a more robust additional input.
Based on the above analysis, we propose a cross-modality domain adaptation framework for freespace detection which exploits both RGB data and surface normal data from source and target domains.
In order to sufficiently employ the multi-modality multi-domain information, our framework considers two essential perspectives for tackling the UDA-FD task, namely inter-modality intra-domain complement and inter-domain intra-modality adaptation.

For the first perspective, it is known that RGB data and surface normal data hold different characteristics. RGB data can well capture color and texture changes in appearance, while surface normal data can well capture height change in 3D space.
In Figure~\ref{fig:intro}(a), we present the freespace predictions of model trained with RGB images only and surface normal maps only respectively.
It is shown that RGB model falsely recognizes shadow areas as non-road regions (box\ding{193} in \textit{RGB Prediction} of Figure~\ref{fig:intro}(a)).
Furthermore, due to the similarity in appearance between roads and sidewalks, it is hard for RGB model to distinguish non-drivable sidewalks from roads as illustrated in box\ding{194} in \textit{RGB Prediction} of Figure~\ref{fig:intro}(a).
The surface normal model, by contrast, classifies these areas into correct labels because it is insensitive to illumination changes and is able to capture height changes in 3D space.
As shown in \textit{Surface Normal Map} of Figure~\ref{fig:intro}(a), a clear boundary can be observed between road and sidewalk areas in the surface normal map, so that it is easy for the model to distinguish them.
On the other hand, as shown in box\ding{192} in \textit{Surface Normal Prediction} of Figure~\ref{fig:intro}(a), for the grassland area where the height changes are minor, surface normal model wrongly labels grassland as road.
However, the RGB model makes correct prediction given to the obvious appearance contrast between road and grassland.
To take advantage of the mutual complement of these two modalities, we devise a Collaborative Cross Guidance (CCG) module which leverages context information from each modality to crossly guide the feature learning of the other modality in both source and target domains.
For each modality, channel-wise global context and spatial-wise response map which is supervised by an auxiliary loss are exploited sequentially to guide the feature learning of the other modality.
In this way, multi-modality information can be sufficiently utilized to correct the flawed predictions collaboratively.

For the second perspective, foreground regions in images from source and target domains contain the same semantic category (\textit{i.e.}, drivable road) with similar style, while background regions usually contain different semantic categories with diverse styles.
For example, in the first column of Figure~\ref{fig:intro}(b), the foreground areas of source and target images are both three marked roads with similar flat ground and gray appearance.
However, the background areas show considerable diversity in both semantic categories and appearances.
The image from source domain contains building and grassland, while the image from target domain contains railway and sidewalks, which has large discrepancy in appearance.
Considering the different characteristics of foreground and background, it is easier and more stable to mainly adapt features of foreground regions between source and target domains.
We accordingly devise a Selective Feature Adaptation (SFA) module which exploits the final predicted logits as attention maps to selectively highlight the foreground region features that are then aligned between source and target domains.
It is noticed that our SFA module only works with RGB images, since there is no obvious difference in style between surface normal maps from the two domains as shown in the second column of Figure~\ref{fig:intro}(b).

Our contributions can be summarized as follows:
\begin{itemize}
    \item To the best of our knowledge, we are the first study to explore unsupervised domain adaptation for freespace detection which assists in alleviating the problem of lacking training data and high annotation cost.
    \item We propose a cross-modality domain adaptation framework utilizing both RGB and surface normal data. A Collaborative Cross Guidance (CCG) module is developed to realize inter-modality intra-domain complement and a Selective Feature Adaptation (SFA) module is developed to realize inter-domain intra-modality adaptation.
    \item Extensive experiments show that our method outperforms other general UDA methods for semantic segmentation and achieves close performance to fully supervised freespace detection methods by adapting the model trained on the source domain to target domain without labels ($93.08\%$ \textit{v.s.} $97.50\%$ F1 score on KITTI-Road dataset~\cite{Fritsch2013ITSC}). 
\end{itemize}

\section{Related Work}

\subsection{Freespace Detection}
Freespace detection is a critical task in autonomous driving, and is beneficial to many other tasks \cite{qi2018dynamic, chen2016dynamically, chen2018parallel}. Traditional methods typically adopt hand-crafted features (e.g., color, edge, texture) of the road and detect the road region with pixel-wise classification~\cite{alvarez2010road,alvarez2013learning,sturgess2009combining}. 
However, the generalization ability of these methods are still far from satisfactory due to their strong reliance on road geometry or appearances.

With the development of deep learning~\cite{huang2020referring, hui2020linguistic, huang2020ordnet} and proposal of Fully Convolutional Network (FCN)~\cite{long2015fully}, methods that regard freespace detection as a semantic segmentation problem have arisen~\cite{lu2019monocular, sun2019reverse, han2018semisupervised, munoz2017deep}. 
For example, Lu~\textit{et al.}~\cite{lu2019monocular} proposes an encoder-decoder architecture to segment road regions on a bird's-eye-viewed image.
Han~\textit{et al.}~\cite{han2018semisupervised} proposes a semi-supervised learning freespace detection method together with a weakly supervised learning method to better leverage the feature distribution of labeled and unlabeled data.
The above RGB-based methods are vulnerable to visual noises like illumination changes and overexposure, which leads to inadequate robustness of these methods. 

To overcome these issues, some methods introduce other modalities of data as supplement of information, such as Lidar~\cite{caltagirone2019lidar, chen2019progressive, gu2021cascaded} and depth~\cite{chang2022fast, fan2020sne, wang2021dynamic, wang2020applying}. 
Gu~\textit{et al.}~\cite{gu2021cascaded} progressively adapts Lidar data space into RGB data space to effectively fuse features of the two modalities.
Chen~\textit{et al.}~\cite{chen2019progressive} proposes to fuse RGB and depth features with a dynamic fusion module (DFM).
Given the assumption that road regions of an image share the same plane, surface normal information estimated from depth images is further incorporated to help with segmenting road regions~\cite{fan2020sne, wang2020applying}.
By including mixture of multi-modality inputs apart from RGB data, the reliability of freespace detection is greatly improved. 

\subsection{Domain Adaptation}

Since directly evaluating on one domain with a model trained on another domain usually causes performance drops, domain adaptation methods are developed to help address the domain shift between two distinct domains. Domain adaptation is widely used in classification~\cite{ghifary2016deep, fang2020dart, wang2021interbn, deng2021informative}, object detection~\cite{chen2018domain, inoue2018cross, wang2019towards, su2020adapting, wang2021exploring}, semantic segmentation~\cite{gao2021dsp, vu2019advent, hoffman2018cycada, zou2018unsupervised, you2021domain} and many other fields~\cite{zhang2019unsupervised, li2021multi, zhang2018facial, zhao2018emotiongan}. 
In particular, unsupervised domain adaptation (UDA) methods have attracted substantial attention since they are free from manually-annotated labels in the target domain. 
UDA methods can be roughly divided into three categories:

\textbf{(1) Feature-level domain adaptation methods.} 
Inspired by the success of generative adversarial network (GAN)~\cite{goodfellow2014generative}, this kind of methods~\cite{tsai2019domain, zheng2019unsupervised, li2020content, huang2020contextual} employs adversarial training scheme to minimize the distance between the feature distribution of source and target domains. 
A discriminator is trained to estimate the domain of latent features generated from the network to make the main prediction.
With the competition, domain-agnostic features will be generated, thereby reducing the domain shift between source and target domains.
Feature-level domain adaptation methods is broadly applied in segmentation tasks~\cite{luo2019taking, vu2019advent, luo2020adversarial, chen2018road, hoffman2016fcns, tsai2018learning} with the assistance with segmentation loss functions.
For example, 
AdaptSeg~\cite{tsai2018learning} applies multi-level adversarial network to align features from different levels, which leads to improved performance. 
ADVENT~\cite{vu2019advent} proposes an entropy-based adversarial learning method, which realizes entropy reduction and feature alignment between domains.

\textbf{(2) Image-level domain adaptation methods.} 
This type of methods performs pixel-level alignment between source and target domains via image-to-image translation strategies such as GAN~\cite{wu2018dcan, hoffman2018cycada, choi2019self, kim2020learning, li2019bidirectional} and style transfer~\cite{yang2020fda, chang2019all, wu2019ace}.
The representative GAN-based methods include CyCADA~\cite{hoffman2018cycada}, which uses CycleGAN~\cite{zhu2017unpaired} to transfer images from source domain into target domain while maintaining semantic consistency with the original images. The synthesized target-styled images share the annotations with original images and can be used as extra training data.
Li~\textit{et al.}~\cite{li2019bidirectional} further proposes to alternately train the image translation network and the segmentation adaptation network to promote each other for better alignment.
Other works like FDA~\cite{yang2020fda} utilizes Fourier transform to stylize images of source domain into target domain and keep the content invariable. 

\textbf{(3) Self-training domain adaptation methods.} 
Self-training is another widely explored type of methods to train the network with the pseudo labels on the target domain to reinforce the network beliefs~\cite{pan2020unsupervised, zhang2021prototypical, huang2020contextual, mei2020instance}. 
Typically, the network is first trained in the source domain, and then used to generate pseudo labels in the target domain, which are subsequently used as supervision to train the target network. After several repeated iterations, the network can also achieve impressive results on the target domain.
In~\cite{li2019bidirectional, zou2018unsupervised, zou2019confidence}, pseudo labels are selected with a confidence threshold calculated based on the whole dataset in an offline manner, which requires storing all predictions for the whole dataset.
To alleviate this issue, Fabio~\textit{et al.}~\cite{pizzati2020domain} propose Weighted Pseudo-Labels (WPL) to estimate a global threshold during optimizing process, reducing the cost of threshold calculation.

\section{Methods}
\subsection{Overview}

\begin{figure*}
  \includegraphics[width=\textwidth]{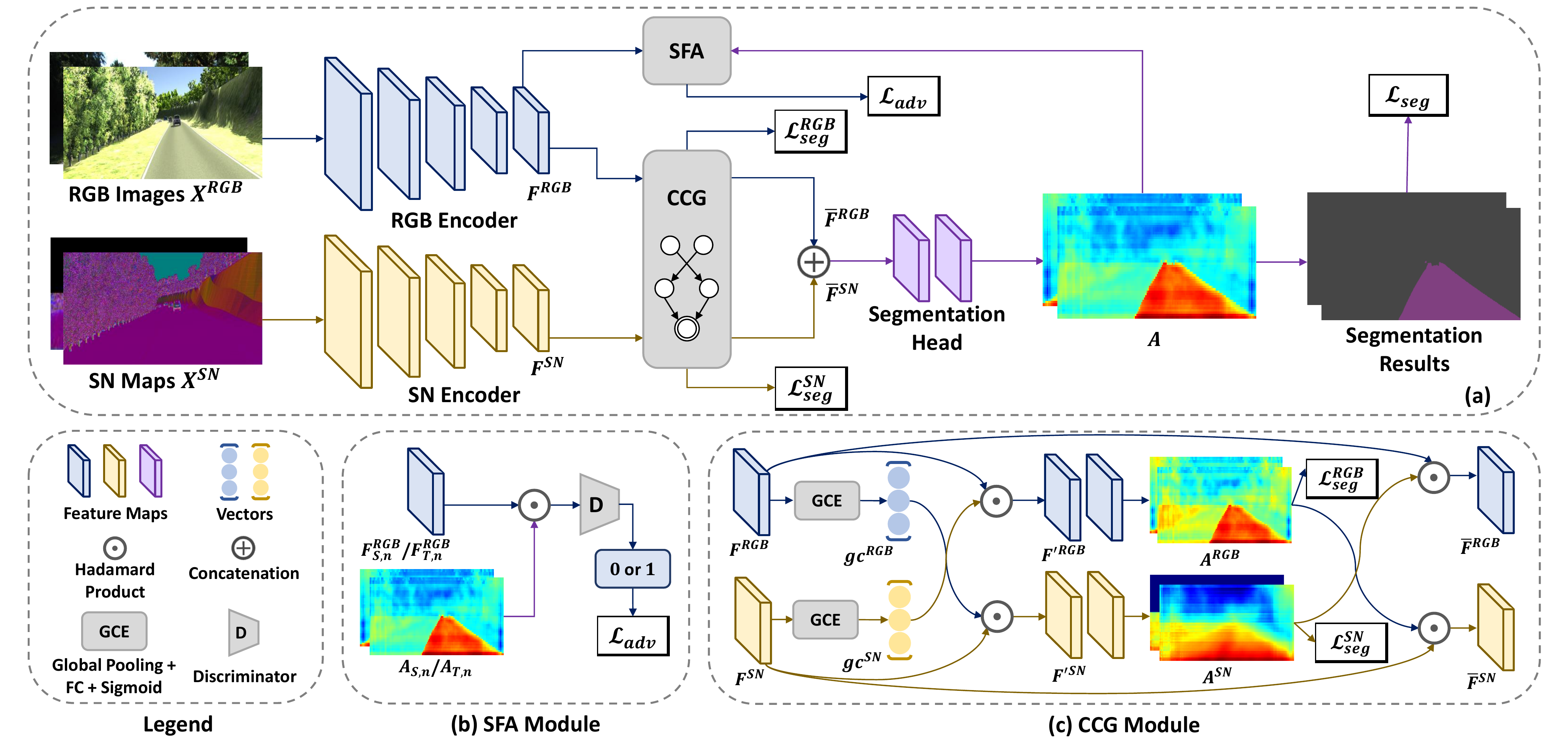}
  \caption{Overall architecture of our proposed method. (a) Pipeline for cross-modality domain adaptation framework. An RGB encoder and a surface normal (SN) encoder are adopted to extract features from source and target domains, followed by CCG module to complement between two modalities. SFA module is leveraged to reduce domain gap of RGB features between source and target domains. (b) Illustration of Selective Feature Alignment (SFA) module. (c) Illustration of Collaborative Cross Guidance (CCG) module.  (Subscripts $S$ and $T$ denoting source and target domains are omitted for simplification in (a) and (c).)}
  \label{fig:pipeline}
\end{figure*}

In this work, we focus on the task of unsupervised domain adaptation for freespace detection, with labels only available for source domain. 
Figure~\ref{fig:pipeline}(a) illustrates the overall architecture of our cross-modality domain adaptation framework.
We take a pair of RGB image and surface normal (SN) map from source or target domain as inputs. The surface normal map is calculated with the corresponding depth image and intrinsic matrix of the camera following~\cite{fan2020sne}.
RGB image $\boldsymbol{X}^{RGB}\in\mathbb{R}^{H \times W \times 3}$ and SN map $\boldsymbol{X}^{SN}\in\mathbb{R}^{H \times W \times 3}$ are processed by an RGB encoder and an SN encoder respectively, where we adopt typical CNN backbones.
The RGB encoder and SN encoder have the same architecture but with different parameters.
The encoded features of the two modalities are denoted as $\boldsymbol{F}^{RGB} \in \mathbb{R}^{H' \times W' \times C}$ and $\boldsymbol{F}^{SN} \in \mathbb{R}^{H' \times W' \times C}$, where $H'$ and $W'$ denote the spatial size of feature map and $C$ denotes the channel number.
Then, $\boldsymbol{F}^{RGB}$ and $\boldsymbol{F}^{SN}$ are fed into the Collaborative Cross Guidance (CCG) module to mutually optimize the feature representation of each other.
In this way, information of multi-modality is collaboratively utilized so as to correct false prediction.
The output feature maps of CCG module are denoted as $\boldsymbol{\bar{F}}^{RGB} \in \mathbb{R}^{H' \times W' \times C}$ and $\boldsymbol{\bar{F}}^{SN} \in \mathbb{R}^{H' \times W' \times C}$ respectively.
Finally, $\boldsymbol{\bar{F}}^{RGB}$ and $\boldsymbol{\bar{F}}^{SN}$ are concatenated to produce the final prediction through a segmentation head.
Since the data from source and target domains share the same network, we omit the domain subscripts $S$ and $T$ for simplification.

In order to reduce the domain shift, we extract $\boldsymbol{F}_{S,n}^{RGB}$ and $\boldsymbol{F}_{T,n}^{RGB}$, which are the $n$-th feature maps of RGB encoder from source and target domains respectively, and apply feature-level alignment between them. 
Together with the foreground attention maps $\boldsymbol{A}_{S}$, $\boldsymbol{A}_T \in \mathbb{R}^{H \times W}$ generated by the segmentation head, $\boldsymbol{F}_{S,n}^{RGB}$ and $\boldsymbol{F}_{T,n}^{RGB}$ are then fed into the Selective Feature Alignment (SFA) module for domain alignment using adversarial strategy.
In SFA module, foreground attention maps are leveraged to selectively align the foreground features between source and target domains, leading to effective bridging of domain gap.

\subsection{Collaborative Cross Guidance Module} 
\label{sec:CCG}
RGB data and SN data have different characteristics. RGB data works well with the obvious texture and color features (\textit{e.g.}, roads \textit{v.s.} grasslands) but fails to distinguish between areas with similar appearance (\textit{e.g.}, roads \textit{v.s.} sidewalks), while SN data works well in capturing height changes but fails to capture texture changes. 
Motivated by the above differences, we introduce a Collaborative Cross Guidance (CCG) module to complement information between the two modalities. 

The detail of CCG module is illustrated in Figure~\ref{fig:pipeline}(c). Feature map $\boldsymbol{F}^{RGB} \in \mathbb{R}^{H' \times W' \times C}$ from RGB encoder and feature map $\boldsymbol{F}^{SN} \in \mathbb{R}^{H' \times W' \times C}$ from SN encoder are taken as inputs of CCG module. To clearly elaborate the process of cross guidance, we take the operations of RGB feature guiding SN feature (denoted as RGB$\rightarrow$SN) as an example. The operations of SN$\rightarrow$RGB are the same.

First of all, the global context vector of SN feature map $\boldsymbol{gc}^{SN} \in \mathbb{R}^{C}$ is extracted to integrate the global information of SN data. Concretely, we first conduct global average pooling on $\boldsymbol{F}^{SN}$ to obtain the global feature of SN data as $\boldsymbol{v}^{SN} \in \mathbb{R}^{C}$. A fully-connected layer together with sigmoid activation is applied to $v^{SN}$ as follow:
\begin{equation}
    \boldsymbol{gc}^{SN} = \sigma(\boldsymbol{W}(\boldsymbol{v}^{SN})),
\end{equation}
where $\sigma(\cdot)$ denotes sigmoid function and $\boldsymbol{W} \in \mathbb{R}^{C \times C}$ is the projection weight. The combination of global average pooling, fully-connected layer and sigmoid activation is abbreviated as $GCE$ in Figure~\ref{fig:pipeline}(c).
The global context vector $\boldsymbol{gc}^{SN}$ is then multiplied to the RGB feature map $\boldsymbol{F}^{RGB}$ to pre-inform the RGB branch with SN information for generating better representations. The generated feature map is denoted as $\boldsymbol{F'}^{RGB} \in \mathbb{R}^{H' \times W' \times C}$.

In order to suppress the mis-response of SN feature with RGB information, we propose to leverage the foreground response map of RGB feature to filter out the false positive responses of SN feature.
To this end, we extract the foreground logit, which is generated with a segmentation head applied to the enhanced RGB feature $\boldsymbol{F'}^{RGB}$, as the foreground attention map of RGB modality.
The RGB foreground attention map $\boldsymbol{A}^{RGB} \in \mathbb{R}^{H' \times W'}$ is obtained as follows:
\begin{gather}
    \label{eq:attn1}
     \boldsymbol{\tilde{P}} = softmax([\boldsymbol{P}_0, \boldsymbol{P}_1]), \\
     \label{eq:attn2}
     \boldsymbol{A}^{RGB} = \boldsymbol{\tilde{P}}_1,
\end{gather}
where $\boldsymbol{P} \in \mathbb{R}^{2 \times H' \times W'}$ is the output of segmentation head, with channel-$0$ denoting the probability of being background and channel-$1$ denoting the probability of being foreground.
The larger value of $\boldsymbol{A}^{RGB}_{i,j}$ indicates the higher confidence of position $(i,j)$ being recognized as foreground.
To ensure that the foreground areas are precisely predicted on $\boldsymbol{A}^{RGB}$, auxiliary segmentation loss $\mathcal{L}^{RGB}_{seg}$ is introduced as supervision. 
Similarly, auxiliary segmentation loss $\mathcal{L}^{SN}_{seg}$ is applied in SN$\rightarrow$RGB process.

Finally, the SN feature map $\boldsymbol{F}^{SN}$ is multiplied with $\boldsymbol{A}^{RGB}$ in an element-wise manner to enhance response of foreground areas and suppress response of background areas, thus eliminating the false positive areas on SN features:
\begin{equation}
    \boldsymbol{\bar{F}}^{SN} = \boldsymbol{F}^{SN} \odot \boldsymbol{A}^{RGB},
\end{equation}
where $\odot$ denotes the Hadamard product.
Through the above interactions between RGB and SN features,  we can leverage useful information of one modality to guide the other and produce better feature representations of both modalities for the final prediction.

\subsection{Selective Feature Alignment Module} 
\label{sec:SFA}
In order to adapt the model trained on the source domain to the target domain, we develop a Selective Feature Alignment (SFA) module for feature-level domain adaptation.
We adopt adversarial strategy to align the features from source and target domains, with a discriminator that distinguishes which domain the feature belongs to. 
Furthermore, as described in Figure~\ref{fig:intro}(b), it is easier to adapt features of only foreground area with the same category than to adapt features of background area with diverse semantics. 
To this end, before feature maps are fed into the discriminator, we first apply foreground attention maps to selectively suppress responses of background areas, so as to mainly align foreground features from two domains.

As illustrated in Figure~\ref{fig:pipeline}(a), to obtain precise estimation of the foreground area, the foreground segmentation logit produced by the segmentation head serves as the foreground attention map $\boldsymbol{A}_{K} \in \mathbb{R}^{H\times W}, K \in \{S, T\}$.
The value at pixel $(i,j)$ means its probability of being road area.
The generation of $\boldsymbol{A}_{K}$ is similar to Equation~\ref{eq:attn1} and~\ref{eq:attn2}.

The detail of SFA module is presented in Figure~\ref{fig:pipeline}(b).
Given feature map $\boldsymbol{F}_{K,n}^{RGB}$, $K \in \{S, T\}$ extracted from the $n$-th stage of RGB encoder, the foreground attention map $\boldsymbol{A}_{K}$ is firstly downsampled to meet the spatial size of the two features to produce $\boldsymbol{A}_{K,n}$.
Afterwards, the two feature maps are multiplied with $\boldsymbol{A}_{S,n}$ and $\boldsymbol{A}_{T,n}$ respectively to relatively suppress responses at background areas, which conversely enhance the foreground response as follows:
\begin{gather}
    \boldsymbol{\bar{F}}_{S,n}^{RGB} = \boldsymbol{F}_{S,n}^{RGB} \odot  \boldsymbol{A}_{S,n}, \\
    \boldsymbol{\bar{F}}_{T,n}^{RGB} = \boldsymbol{F}_{T,n}^{RGB} \odot \boldsymbol{A}_{T,n}.
\end{gather}
The enhanced feature maps $\boldsymbol{\bar{F}}_{S,n}^{RGB}$ and $\boldsymbol{\bar{F}}_{T,n}^{RGB}$ are then fed to a fully convolutional discriminator $D$ that performs binary classification to determine which domain the feature is from.
In our work, we define label $0$ for source domain and label $1$ for target domain. 
The corresponding loss function is as follows:
\begin{equation}
    \mathcal{L}_{adv} = \sum_{i,j}\log(D(\boldsymbol{\bar{F}}_{T,n}^{RGB})_{i,j})+\log(1-D(\boldsymbol{\bar{F}}_{S,n}^{RGB})_{i,j}),
\end{equation}
where $i, j$ denote the row index and column index respectively.

In the mean time, the RGB-encoder is trained to fool the discriminator by producing feature maps from target domain which are highly similar to feature maps from source domain.
By maximizing the probability of the target prediction being considered as the source prediction, the feature distribution shift between two domains is eliminated in the RGB encoder, thus greatly alleviating the performance degradation on target domain.

The SFA module is only applied to RGB features since surface normal data is domain invariant as shown in Figure~\ref{fig:intro}(b).
Experiments in Section~\ref{sec:abl_study} further demonstrate this hypothesis, so that it is unnecessary to align the feature distribution of surface normal.


\begin{table*}[htbp]
    \centering
    \caption{Freespace detection performance on KITTI-Road validation set with Virtual KITTI, R2D and SYNTHIA-SF utilized as source datasets respectively. ST means self-training with pseudo labels on the target domain.}
    \label{tab:compare_sota}
    \adjustbox{max width=1\linewidth}{
    \begin{tabular}{l|c|c||cccc|cccc|cccc}
    \Xhline{1pt}
    \rowcolor{mygray} &  &  & \multicolumn{4}{c|}{\bf Virtual KITTI $\rightarrow$ KITTI-Road} & \multicolumn{4}{c|}{\bf R2D $\rightarrow$ KITTI-Road} & \multicolumn{4}{c}{\bf SYNTHIA-SF $\rightarrow$ KITTI-Road} \\
    \rowcolor{mygray} \multicolumn{1}{c|}{\multirow{-2}{*}{\bf Method}} & \multirow{-2}{*}{\bf Backbone} & \multirow{-2}{*}{\bf Depth} & \textbf{PRE} & \textbf{REC} & \textbf{F1 score} & \textbf{IoU} & \textbf{PRE} & \textbf{REC} & \textbf{F1 score} & \textbf{IoU} & \textbf{PRE} & \textbf{REC} & \textbf{F1 score} & \textbf{IoU} \\
    \hline\hline
    \multirow{3}{*}{Fully Supervised} & ResNet-101~\cite{He_2016_CVPR} & $\surd$ & 96.79 & 95.51 & 96.15 & 92.78 & 96.79 & 95.51 & 96.15 & 92.78 & 96.79 & 95.51 & 96.15 & 92.78 \\
     & ResNet-50~\cite{He_2016_CVPR} & $\surd$ & 95.04 & 94.92 & 94.98 & 90.44 & 95.04 & 94.92 & 94.98 & 90.44 & 95.04 & 94.92 & 94.98 & 90.44 \\
     & HRNetV2-W48~\cite{hrnet} & $\surd$ & 96.83 & 96.28 & 96.55 & 93.33 & 96.83 & 96.28 & 96.55 & 93.33 & 96.83 & 96.28 & 96.55 & 93.33 \\
    \hline
    w/o adaptation & \multirow{5}{*}{ResNet-101~\cite{He_2016_CVPR}} & & 84.04 & 90.61 & 87.18 & 77.28 & 49.58 & 98.65 & 65.99 & 49.25 & 56.47 & 39.29 & 46.34 & 30.16 \\
    AdaptSeg \cite{tsai2018learning} &  & & 85.22 & 95.02 & 89.85 & 81.58 & 52.71 & \textbf{98.83} & 68.75 & 52.38 & 59.80 & 97.56 & 74.15 & 58.92 \\
    ADVENT \cite{vu2019advent} &  & & 83.48 & \textbf{97.35} & 89.88 & 81.62 & 54.24 & 98.04 & 69.84 & 53.65 & 63.14 & 95.00 & 75.86 & 61.11 \\
    IntraDA \cite{pan2020unsupervised} &  & & 86.29 & 96.89 & 91.28 & 83.96 & 58.98 & 97.53 & 73.51 & 58.11 & 64.30 & \textbf{97.69} & 77.55 & 63.33 \\
    DADA \cite{vu2019dada} &  & $\surd$ & 86.02 & 95.14 & 90.35 & 82.40 & 53.67 & 98.70 & 69.53 & 53.29 & 57.62 & 97.28 & 72.38 & 56.71 \\
    \hline
    \textbf{Ours (w/o ST)} & \multirow{2}{*}{ResNet-50~\cite{He_2016_CVPR}} & $\surd$ & 91.28 & 94.36 & 92.79 & 86.56 & 78.78 & 95.31 & 86.26 & 75.84 & 77.09 & 87.47 & 81.95 & 69.43 \\
    \textbf{Ours (w/ ST)} &  & $\surd$ & \textbf{93.41} & 94.53 & 93.97 & 88.62 & \textbf{86.27} & 92.84 & \textbf{89.43} & \textbf{80.89} & 78.36 & 94.99 & 85.88 & 75.25 \\
    \cline{2-2}
    \textbf{Ours (w/o ST)} & \multirow{2}{*}{HRNetV2-W48~\cite{hrnet}} & $\surd$ & 90.73 & 96.09 & 93.33 & 87.5 & 73.78 & 98.04 & 83.75 & 72.04 & 76.12 & 89.88 & 82.43 & 70.11 \\
    \textbf{Ours (w/ ST)} &  & $\surd$ & 91.88 & 97.00 & \textbf{94.37} & \textbf{89.34} & 82.32 & 97.39 & 89.22 & 80.54 & \textbf{85.55} & 94.67 & \textbf{89.88} & \textbf{81.62} \\
    \bottomrule
    \end{tabular}
    }

\end{table*}

\subsection{Training Scheme} 
\label{sec:training}
We use self-training method to train our model in an iterative manner.
The training procedure of our method contains $3$ rounds.
In the first round, since the labels of target domain are not available, we only calculate the segmentation loss for the source domain. Thus, the complete loss function for round $1$ is designed as:
\begin{equation}
    \mathcal{L} = \lambda_{1,S}\mathcal{L} ^{RGB}_{seg,S} + \lambda_{2,S}\mathcal{L} ^{SN}_{seg,S} + \lambda_{3,S}\mathcal{L}_{seg,S} + \lambda_{4}\mathcal{L}_{adv},
\end{equation}
where $\lambda_{(\cdot)}$ denotes the weight of each term. 
For the last $2$ rounds of training, we generate pseudo labels for target domain with the network trained in the previous round. 
Concretely, we set a threshold $\alpha$ to ignore low-confidence pixels with classification probability between $1-\alpha$ and $\alpha$ when calculating segmentation losses.
Thus, the complete loss function for the last $2$ rounds is defined as:
\begin{equation}
\begin{aligned}
    \mathcal{L} = &  \lambda_{1,S}\mathcal{L}^{RGB}_{seg,S} + \lambda_{2,S}\mathcal{L}^{SN}_{seg,S} + \lambda_{3,S}\mathcal{L}_{seg,S} 
    + \lambda_{1,T}\mathcal{L}^{RGB}_{seg,T} \\
    & + \lambda_{2,T}\mathcal{L}^{SN}_{seg,T} + \lambda_{3,T}\mathcal{L}_{seg,T} +\lambda_{4}\mathcal{L}_{adv}.
\end{aligned}
\end{equation}
With supervision on the target domain, the performance can be further improved.
The ultimate objective of our method is optimized with a min-max strategy:
\begin{equation}
    \mathop{min}\limits_{\boldsymbol{G}}\mathop{max}\limits_{\boldsymbol{D}}(\mathcal{L}),
\end{equation}
where $\boldsymbol{G}$ denotes the segmentation network.
By alternately training the segmentation network and the discriminator, the segmentation losses of both domains are minimized, while the probability of target feature being regarded as source feature is maximized.

\section{Experiments}

\subsection{Datasets and Evaluation Metrics}

In the experiment part, we adopt the setting of adaptation from synthetic datasets to real-world datasets.
Three synthetic datasets including R2D~\cite{fan2020sne}, SYNTHIA-SF~\cite{hernandez2017slanted} and Virtual KITTI~\cite{gaidon2016virtual} are used as source domains and one real-world dataset KITTI-Road~\cite{Fritsch2013ITSC} is used as target domain.
Detailed description on these datasets can be found in Appendix~\ref{sec: datasets}.
Since multiple submission on the test set of KITTI-Road dataset is not supported, we follow most of the previous works~\cite{fan2020sne,chang2022fast} to divide the original training set into $173$ for training and $116$ for validation. Unless otherwise specified, the experiments following are conducted on these splits.

We adopt the most commonly used metrics including precision (PRE), recall (REC), F1 score and the intersection over union (IoU) as our evaluation metrics. 
Their corresponding definitions are as follows: Precision = $\frac{TP}{TP+FP}$, Recall = $\frac{TP}{TP+FN}$, F1-score = $\frac{2\times Precision \times Recall}{Precision + Recall}$ and IoU = $\frac{TP}{TP+FP+FN}$. $TP, FP$ and $FN$ represent the true positive, false positive and false negative pixel numbers. 
Taking $TP$, $TN$, $FP$ and $FN$ as the number of true positive, true negative, false positive and false negative predicted pixels respectively, these metrics are defined as follows: 
PRE = $\frac{TP}{TP+FP}$, REC = $\frac{TP}{TP+FN}$, F1 score = $\frac{2\times PRE \times REC}{PRE + REC}$ and IoU = $\frac{TP}{TP+FP+FN}$.

\subsection{Implementation Details}
Two candidates for RGB/SN encoders are adopted in our work including ResNet-50~\cite{He_2016_CVPR} and HRNetV2-W48~\cite{hrnet}.
Both networks are pretrained on ImageNet dataset~\cite{imagenet}. 
The segmentation heads of the main decoder and auxiliary decoders in CCG module all contain two $1\times 1$ convolutions.
The discriminator consists of four strided convolutional layers for feature map downsampling and LeakyReLU~\cite{xu2015empirical} is adopted as the activation function with negative slope set to $0.2$. 
For experiments with ResNet-50, only the last stage of feature maps produced by the two encoders are included for following operations. $1\times 1$ convolutions are applied first to reduce the channel number from $2048$ to $512$.
For experiments with HRNet, $4$ feature maps with hierarchical resolutions are produced by the encoder for each modality.
In CCG module, only the largest RGB/SN feature maps among them are used to generate foreground attention maps $\boldsymbol{A}^{RGB}$ and $\boldsymbol{A}^{SN}$. Afterwards, all the encoder feature maps of both modalities are multiplied with the corresponding foreground attention maps for cross-modality guidance and then stacked together before feeding to segmentation head for final prediction.
As for the SFA module, feature-level alignment is only preformed on the largest feature map.

Our method is implemented using PyTorch~\cite{paszke2017automatic} and experiments are conducted on $4$ NVIDIA GTX 1080Ti GPUs with 11GB memory.
To train the segmentation network, we apply an SGD optimizer with a learning rate of $2.5\times 10^{-4}$, momentum of $0.9$ and weight decay of $5 \times 10^{-4}$. Adam~\cite{kingma2014adam} with default parameters is utilized as the optimizer of the discriminator. As for the weight of each loss, we set $\lambda_{1,S}=\lambda_{2,S}=0.5$, $\lambda_{3,S}=1.0$, $\lambda_{1,T}=\lambda_{2,T}=0.2$, $\lambda_{3,T}=0.5$, $\lambda_{4}=10^{-3}$ for experiments with HRNetV2-W48 and $\lambda_{4}=10^{-4}$ for experiments with ResNet-50. The threshold $\alpha$ for generating pseudo labels is set to $0.99$.

\subsection{Comparison with State-of-the-art Methods} 
\label{compare_sota}
To demonstrate the effectiveness of our proposed framework, we carry out experiments to adapt from $3$ different synthetic datasets to the real-world dataset KITTI-Road respectively. The results on the validation set are reported in Table~\ref{tab:compare_sota}.
Since we are the first to explore the UDA-FD task, there is no previous works in this area to compare with.
Thus we pick out several methods of UDA for semantic segmentation which could be adapted to our task and reproduce these methods based on their released codes.
The selected methods includes AdaptSeg~\cite{tsai2018learning}, which is based on feature-level adaptation, ADVENT~\cite{vu2019advent} and IntraDA~\cite{pan2020unsupervised}, which seek to minimize the entropy map, and DADA~\cite{vu2019dada}, which incorporates depth images as extra supervision. 
We also conduct more experiments using both RGB and depth information to demonstrate the superiority of our method in Appendix~\ref{sec: more_exp}.
As shown in the first $3$ rows of Table~\ref{tab:compare_sota}, we train our segmentation network with labeled images on KITTI-Road to provide an upper bound of KITTI-Road validation set.

\begin{table}[tbp]
    \centering
    \caption{Comparison among our method, other freespace detection methods and other UDA methods for semantic segmentation on the KITTI-Road benchmark.}
    \label{tab:other_freespace}
\adjustbox{max width=\linewidth}{
    \begin{tabular}{c|c||c|c|c|c|c}
        \Xhline{1pt}
        \rowcolor{mygray} \textbf{Setting} & \textbf{Method} & \textbf{MaxF} & \textbf{AP} & \textbf{PRE} & \textbf{REC} & \textbf{Rank} \\
        \hline\hline
        \multirow{3}{*}{Fully supervised} & SNE-RoadSeg+ \cite{wang2021sne} & 97.50 & 93.98 & 97.41 & 97.58 & 2 \\
          & USNet~\cite{chang2022fast} & 96.89 & 93.25 & 96.51 & 97.27 & 5 \\
          & NIM-RTFNet~\cite{wang2020applying} & 96.02 & 94.01 & 96.43 & 95.62 & 12 \\
        \hline
        \multirow{2}{*}{Semi-Supervised} & SSLGAN~\cite{han2018semisupervised} & 95.53 & 90.35 & 95.84 & 95.24 & 17 \\
          & RGB36-Super~\cite{caltagirone2019lidar} & 92.94 & 92.29 & 93.14 & 92.74 & 39 \\
        \hline
        \multirow{2}{*}{Unsupervised} & ADVENT~\cite{vu2019advent} & 89.61 & 89.34 & 87.96 & 91.31 & 52 \\
        \multirow{2}{*}{Domain Adaptation} & DADA~\cite{vu2019dada} & 89.71 & 89.35 & 88.44 & 91.02 & 50 \\
          & \textbf{Ours} & \textbf{93.08} & \textbf{93.00} & \textbf{93.19} & \textbf{92.96} & 38 \\ 
        \bottomrule
    \end{tabular}
}
\end{table}

\begin{figure*}[!htb]
  \includegraphics[width=\textwidth]{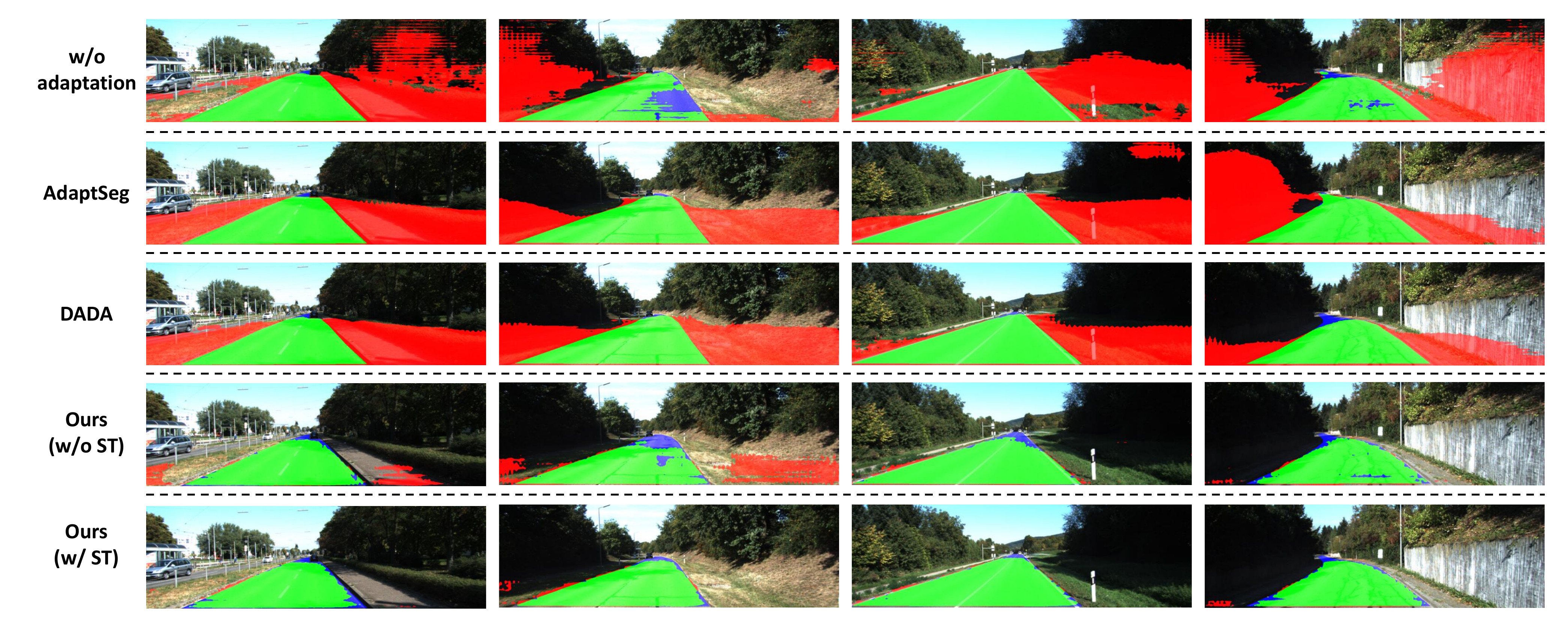}
  \caption{Qualitative comparison between our model and other methods on R2D$\rightarrow$KITTI-Road. The true positive, false negative and false positive pixels are respectively shown in \textcolor{green}{green}, \textcolor{blue}{blue} and \textcolor{red}{red}.}
  \label{fig:vis_sota}
\end{figure*}

\begin{figure*}[]
  \includegraphics[width=\textwidth]{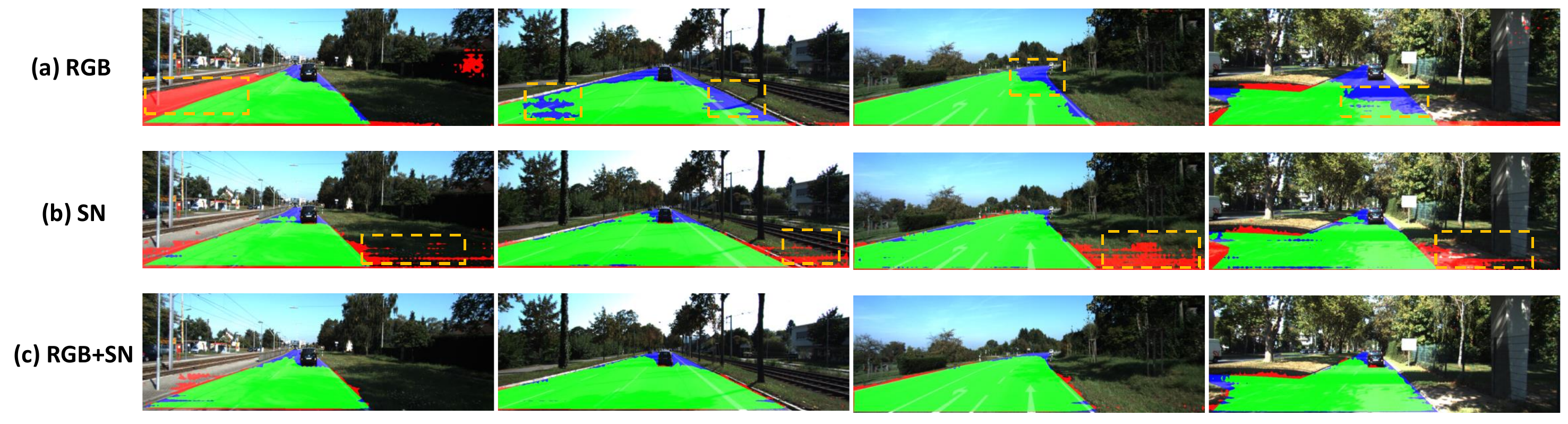}
  \caption{Qualitative results of models using different modalities. The true positive, false negative and false positive pixels are respectively shown in \textcolor{green}{green}, \textcolor{blue}{blue} and \textcolor{red}{red}. The regions where RGB or SN model makes false predictions are highlighted within yellow dashed boxes. (a) Results of RGB model. (b) Results of SN model. (c) Results of our full model.}
  \label{fig:vis_modal}
\end{figure*}

Take Virtual KITTI$\rightarrow$KITTI-Road as an example, our method with ResNet-50 as backbone achieves high performances on the most important metrics including F1 score and IoU compared with the other UDA segmentation methods such as DADA, which also leverages depth images.
$1.18\%$ and $2.06\%$ performance gains of F1 score and IoU are obtained by training with pseudo labels (denoted as \textit{w/ ST}).
Replacing the backbone with HRNetV2-W48 further boosts the performance to $94.37\%$ F1 score and $89.34\%$ IoU, which is only $2.18\%$ lower than full-supervision baseline on F1 score, demonstrating the potential of unsupervised domain adaptation framework for freespace detection.
It is interesting that other UDA methods like ADVENT~\cite{vu2019advent} and IntraDA~\cite{pan2020unsupervised} typically achieve high recall but much lower precision and IoU, implying that these methods tend to falsely include confusing background areas into foreground areas for lacking complement between RGB and SN information.
The visualization results shown in Figure~\ref{fig:vis_sota} further prove it.
Similar performance gap could be observed on R2D$\rightarrow$KITTI-Road and SYNTHIA-SF$\rightarrow$KITTI-Road except that the overall performances are lower due to the larger domain gap of these two settings, which indicates that the gap magnitude between source and target domains is one of the key factors to affect the adaptation performance.

In addition, we compare with other published fully-supervised and semi-supervised methods on KITTI-Road Benchmark and the results are presented in Table~\ref{tab:other_freespace}.
We use Virtual KITTI as source dataset and the original training set of KITTI-Road as target dataset.
It is shown that our method outperforms all the UDA methods and one of the semi-supervised methods RGB36-super~\cite{caltagirone2019lidar}. Besides, our method achieves close AP and MaxF to the fully supervised methods, further indicating the large potential of our method.

\subsection{Ablation Study} 
\label{sec:abl_study}
To evaluate different design of our model, we conduct ablation study on R2D$\rightarrow$KITTI-Road with ResNet-50 as backbone. 

\textbf{Component analysis.} 
Table~\ref{tab:ablation} presents the ablation results of input modalities and our proposed modules. 
The $1$-st row denotes the RGB-only results without any adaptation to the target domain, which are far from satisfactory. 
By inserting the SFA module to reduce domain gap ($2$-nd row), notable gain with $18.25\%$ on F1 score and $22.10\%$ on IoU is obtained, demonstrating the necessity for feature alignment in domain adaptation.
In the $3$-rd row, surface normal information is introduced as supplement of RGB information and brings $5.03\%$ F1 score and $7.28\%$ IoU improvements, showing the effectiveness of multi-modal learning. Finally, together with CCG module to complement between RGB and SN data, the performances are further boosted, resulting in the highest scores on both F1 score and IoU. 

\textbf{Feature modalities to adapt in SFA.} 
In Table~\ref{tab:abl_sfa}, we evaluate different modalities for feature alignment in SFA module. 
The $1$-st row shows the results of simply involving two modalities without adaptation.
By aligning only RGB features from both domains, notable performance gain ($5.53\%$ improvement in IoU) is obtained in the $2$-nd row.
However, when applying feature alignment to SN features in the $3$-rd row, obvious performance drop could be observed comparing with the $1$-st row.
Aligning both RGB and SN features achieves sub-optimal performances comparing with aligning RGB features only as shown in the $4$-th row.
The above results indicate that SN data is domain-invariant to some extent. In our work, we only adapt the feature distribution of RGB data.


\begin{table}[]
\centering
    \caption{Ablation study to verify effectiveness of each component on R2D$\rightarrow$KITTI-Road with ResNet-50 as backbone.}
    \label{tab:ablation}
\adjustbox{max width=\linewidth}{
        \begin{tabular}{r|cccc}
            \Xhline{1pt}
            \rowcolor{mygray} \textbf{Settings} & \textbf{F1 score} & \textbf{$\Delta$F1} & \textbf{IoU} & \textbf{$\Delta$IoU} \\
            \hline\hline
            RGB & 61.69 & - & 44.49 & - \\
            +SFA & 79.94 & 18.25 & 66.59 & 22.10 \\
            +SN & 84.97 & 5.03 & 73.87 & 7.28 \\
            +CCG & \textbf{86.26} & 1.29 & \textbf{75.84} & 1.97 \\
            \bottomrule
        \end{tabular}
}
\end{table}

\begin{table}[]
    \centering
    \caption{Ablation study of SFA module being applied to different modalities. SFA-RGB means adapting RGB features in SFA module and SFA-SN means adapting SN features in SFA module.}
    \label{tab:abl_sfa}
\adjustbox{max width=\linewidth}{
    \begin{tabular}{cc|cc}
        \Xhline{1pt}
        \rowcolor{mygray} \textbf{SFA-RGB} & \textbf{SFA-SN} & \textbf{F1 score} & \textbf{IoU} \\
        \hline\hline
          & & 81.19 & 68.34 \\
        $\surd$ & & \textbf{84.97} & \textbf{73.87} \\
         & $\surd$ & 79.06 & 65.37 \\
        $\surd$ & $\surd$ & 81.84 & 69.27 \\
        \bottomrule
    \end{tabular}
}
\end{table}

\subsection{Qualitative Analysis}

As shown in Figure~\ref{fig:vis_sota}, we visualize the predictions of our method and others on the R2D$\rightarrow$KITTI-Road setting.
For example, in the $4$-th column, AdaptSeg~\cite{tsai2018learning} and DADA~\cite{vu2019dada} predict sidewalk on the right as road area, while our method is able to differentiate these two categories. Although depth information is also utilized in DADA~\cite{vu2019dada}, it is not sufficient to suppress the false predictions on sidewalks.
In addition, by training with the selected high-confidence pixels as supervision, our method generates predictions with cleaner boundaries as shown in the $4$-th row.
Our method shows superiority in distinguishing road area and other road-like areas compared with other UDA methods.
Qualitative results on the other two settings are shown in Appendix~\ref{sec: more_vis}.

We also visualize the segmentation results predicted with different modalities in Figure~\ref{fig:vis_modal}. Mutual complement between RGB and SN data can be observed from their predictions.
For example, in the $1$-st column, SN model can roughly distinguish the sidewalk and road, while RGB model mis-classifies the sidewalk totally.
In the $3$-rd column, SN model fails to distinguish between the grassland and road, while the RGB model makes the correct prediction on grassland area.
By fusing the two modalities, wrong predictions of one modality could be corrected by the other, thus generating precise segmentation results.

\section{Conclusion}
In this paper, we explore unsupervised domain adaptation for free-space detection (UDA-FD) between synthetic and real-world data to resolve the data limitation problem and alleviate annotation costs. To the best of our knowledge, we are the first to investigate the UDA-FD task. We propose a cross-modality domain adaptation framework which simultaneously considers both inter-modality intra-domain complement and
inter-domain intra-modality adaptation. Extensive experimental results show that our proposed method achieves close performance to fully supervised freespace detection methods, demonstrating the potential of UDA for freespace detection.
In the future, we hope to explore more domain adaptation techniques for freespace detection, and generalize our cross-modality domain adaptation framework to other settings, such as freespace detection under extreme weather conditions.

\begin{acks}
This work was supported in part by the National Natural Science Foundation of China under Grant $62122010$ and Grant $61876177$, in part by the Fundamental Research Funds for the Central Universities, and in part by the Key Research and Development Program of Zhejiang Province under Grant $2022$C$01082$. 
\end{acks}

\balance
\normalem
\bibliographystyle{ACM-Reference-Format}
\bibliography{ref}

\newpage

\nobalance

\appendix

\section{Unique Points of the UDA-FD Task} \label{sec: unique_fd}

It is true that UDA-FD shares some similarity with UDA for semantic segmentation (UDA-SS), with both two tasks aim at pixel-level classification. However, there are still difference between them, which lies mainly in two aspects:

The UDA-SS task labels almost each pixel with one of a closed set of known categories. However, UDA-FD task separates foreground pixels (\textit{i.e.}, drivable area) from background pixels which are composed of an open set of multiple unknown categories. Therefore, the uncertainty of background semantics leads to the difficulty to adapt between source and target domains. To alleviate this issue, we propose the SFA module which aligns only foreground features to diminish the disturbance of complex background feature. Besides, the uncertainty of background semantics also prevents UDA-SS methods based on minimizing entropy to perform well on the UDA-FD task.
    
In UDA-FD task, there are strong appearance similarity between foreground and some background categories such as sidewalk and curb, which are hard to distinguish from the foreground drivable area only based on RGB visual appearance. As a result, geometric properties, \textit{e.g.} the planar properties of the road, are usually important cues in UDA-FD task to distinguish these confusing categories. In our method, we integrate the geometric surface normal information into our framework as a complement with RGB information in CCG module. As for UDA-SS task, such properties are relatively less explored since the confusing categories, such as sidewalk and curb, also have their own pixel-level annotations from which UDA-SS methods can directly learn.

\section{Datasets} \label{sec: datasets}
\textbf{Ready-to-Drive (R2D)}~\cite{fan2020sne} is a large-scale synthetic freespace detection dataset collected with the CARLA~\cite{dosovitskiy2017carla} simulator. 
It consists of $6$ different scenarios with $11,430$ pairs of stereo images, together with their corresponding depth images and segmentation labels. 

\textbf{SYNTHIA-SF}~\cite{hernandez2017slanted} is a synthetic semantic segmentation dataset consisting of $6$ video sequences under different scenarios and traffic conditions. 
$2,224$ images along with depth and segmentation labels are included in total.
We combine the road and road lines classes to form the foreground areas while others are treated as background. 
All of the $2,224$ images are used for training.

\textbf{Virtual KITTI}~\cite{gaidon2016virtual} is a synthetic dataset generated from five different virtual worlds in urban settings under different imaging and weather conditions.
There are $34$ high-resolution videos together with accurate ground truth for depth and segmentation labels available in this dataset.
In the experiment, we sample a total number of $2,126$ frames for training. 

\textbf{KITTI-Road}~\cite{Fritsch2013ITSC} is a real-world dataset for freespace detection.
It contains $579$ images in total with $289$ for training and $290$ for testing.
We regard road class as foreground and the others as background following~\cite{fan2020sne}.

\begin{table}[htbp]
    \centering
    \caption{Freespace detection performance on KITTI-Road validation set with R2D utilized as source datasets. We integrate depth information in RGB-only methods by concatenating depth map with RGB image as input to the model.}
    \label{tab: more_rgbd_results}
    \adjustbox{max width=1\linewidth}{
    \begin{tabular}{l|c||cccc}
    \Xhline{1pt}
    \rowcolor{mygray} \multicolumn{1}{c|}{\textbf{Method}} & \textbf{Depth} & \textbf{PRE} & \textbf{REC} & \textbf{F1 score} & \textbf{IoU} \\
    \hline\hline
    AdaptSeg \cite{tsai2018learning} & & 52.71 & 98.83 & 68.75 & 52.38 \\
    AdaptSeg \cite{tsai2018learning} & $\surd$ & 59.87 & \textbf{98.93} & 74.60 & 59.48 \\
    ADVENT \cite{vu2019advent} & & 54.24 & 98.04 & 69.84 & 53.65 \\
    ADVENT \cite{vu2019advent} & $\surd$ & 56.70 & 98.32 & 71.92 & 56.15 \\
    IntraDA \cite{pan2020unsupervised} & & 58.98 & 97.53 & 73.51 & 58.11 \\
    IntraDA \cite{pan2020unsupervised} & $\surd$ & 60.38 & 96.44 & 75.23 & 60.53 \\
    \textbf{Ours-ResNet50} & $\surd$ & \textbf{86.27} & 92.84 & \textbf{89.43} & \textbf{80.89} \\
    \bottomrule
    \end{tabular}
    }
\end{table}

\begin{table}[htbp]
    \centering
    \caption{Comparison with more recent works on R2D$\rightarrow$KITTI-Road setting.}
    \label{tab: more_sota}
\adjustbox{max width=\linewidth}{
    \begin{tabular}{c|c||c|c|c|c}
        \Xhline{1pt}
        \rowcolor{mygray} \textbf{Method} & \textbf{Publication} & \textbf{PRE} & \textbf{REC} & \textbf{F1 Score} & \textbf{IoU} \\
        \hline\hline
        PCLA~\cite{kang2020pixel} & NeurIPS 2020 & 57.74 & \textbf{95.88} & 72.07 & 56.34 \\
        CCM~\cite{li2020content} & ECCV 2020 & 63.25 & 94.58 & 75.80 & 61.04 \\
        Ours-ResNet50 & - & \textbf{86.27} & 92.84 & \textbf{89.43} & \textbf{80.89} \\
        \bottomrule
    \end{tabular}
}
\end{table}

\section{Experiment Results} \label{sec: more_exp}
Since there is only DADA~\cite{vu2019dada} that uses depth modality in UDA-SS task, we integrate depth modality into other RGB-only methods by concatenating the depth feature with RGB feature extracted by seperated encoders for comparison with our method. Results on R2D$\rightarrow$KITTI-Road dataset are summarized in Table~\ref{tab: more_rgbd_results}. As shown in the table, incorporating depth information to these UDA-SS methods can obviously improve their performance, but our method still outperforms these ones by large margins on F1 Score and IoU.

To further validate the effectiveness of our method, we adapt more recent works including PCLA~\cite{kang2020pixel} and CCM~\cite{li2020content} to UDA-FD task and report the results on R2D -> KITTI-ROAD with ResNet-101 as backbone in Table~\ref{tab: more_sota}. Our method outperforms the two works in terms of both F1 Score and IoU.

\section{Qualitative Results} \label{sec: more_vis}

\begin{figure*}[!htb]
  \includegraphics[width=\textwidth]{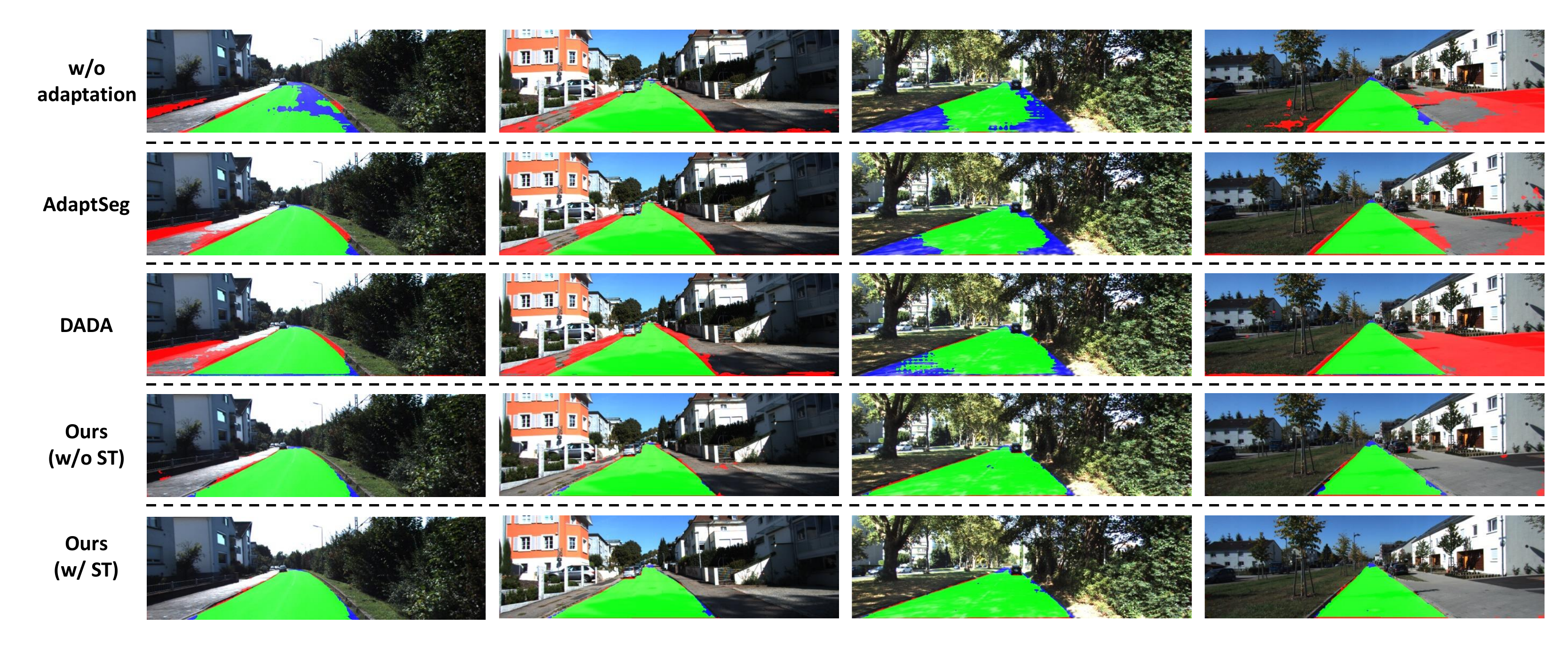}
  \caption{Qualitative comparison between our model and other methods on Virtual KITTI$\rightarrow$KITTI-Road. The true positive, false negative and false positive pixels are respectively shown in \textcolor{green}{green}, \textcolor{blue}{blue} and \textcolor{red}{red}.}
  \label{fig:vis_vkitti}
\end{figure*}

\begin{figure*}[!htb]
  \includegraphics[width=\textwidth]{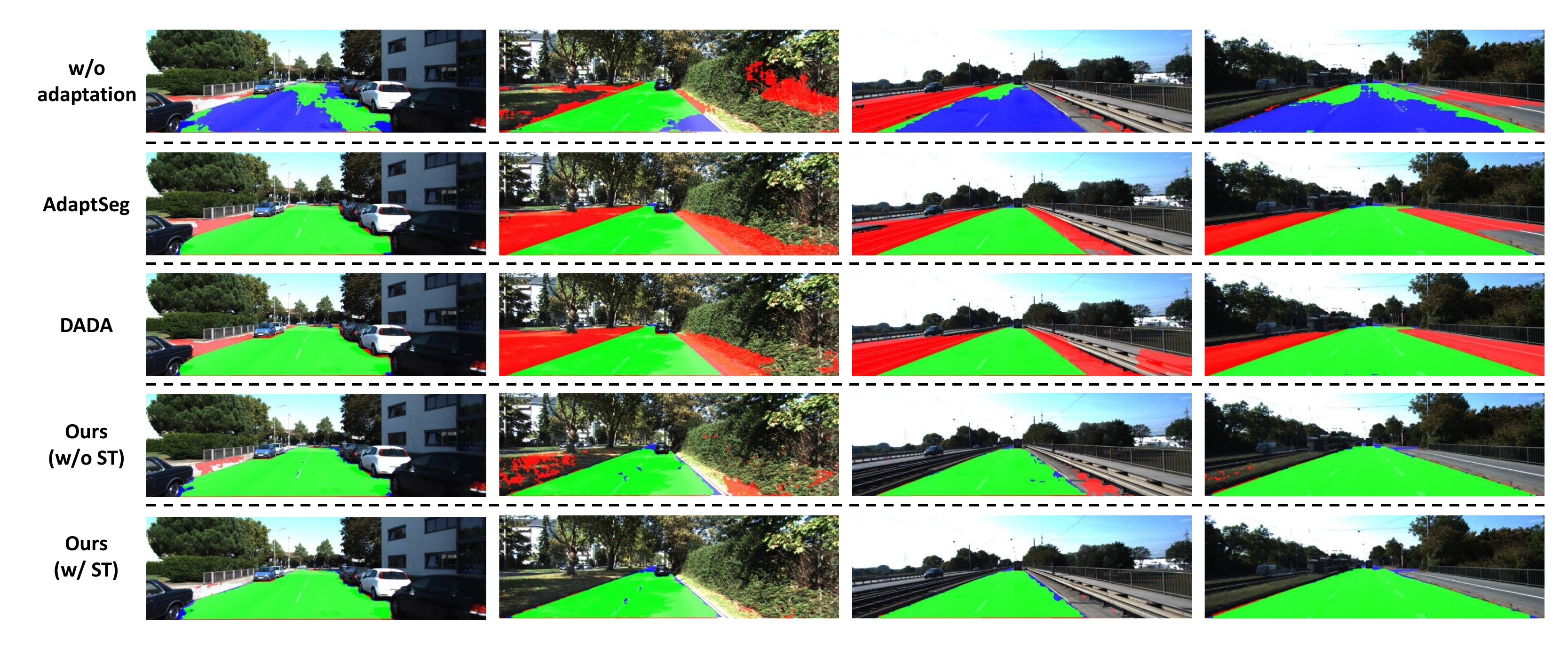}
  \caption{Qualitative comparison between our model and other methods on SYNTHIA-SF$\rightarrow$KITTI-Road. The true positive, false negative and false positive pixels are respectively shown in \textcolor{green}{green}, \textcolor{blue}{blue} and \textcolor{red}{red}.}
  \label{fig:vis_syn}
\end{figure*}

In Figure~\ref{fig:vis_vkitti} and Figure~\ref{fig:vis_syn}, we visualize qualitative results of our model and other methods on Virtual KITTI $\rightarrow$ KITTI-Road and SYNTHIA-SF $\rightarrow$ KITTI-Road settings respectively.
We can observe that AdaptSeg~\cite{tsai2018learning} and DADA~\cite{vu2019dada} fail to distinguish sidewalks or grasslands from roads (\textit{e.g.}, $4$-th column in Figure~\ref{fig:vis_vkitti} and $2$-nd column in Figure~\ref{fig:vis_syn}) and cannot completely segment the road area when there 
are shadows (\textit{e.g.}, $3$-rd column in Figure~\ref{fig:vis_vkitti}). However, in the predictions of our method, there are much less mis-classified pixels in these regions, demonstrating the effectiveness of selective feature alignment and cross-modal guidance. As shown in the last row of Figure~\ref{fig:vis_vkitti} and Figure~\ref{fig:vis_syn}, with the help of self-training, most false predictions on sidewalks are suppressed.

\end{document}